\documentclass[10pt,twocolumn,letterpaper]{article}

\usepackage{cvpr}
\usepackage{times}
\usepackage{epsfig}
\usepackage{graphicx}
\usepackage{amsmath}
\usepackage{amssymb}
\usepackage{dsfont}
\usepackage{siunitx}
\usepackage{subfigure}
\usepackage{multirow}
\usepackage{caption}
\usepackage{booktabs}       %
\usepackage{standalone}
\usepackage{xspace}
\usepackage{comment}
\usepackage{multirow}

\graphicspath{{./images_lowres/}}

\def\1{\mathds{1}}

 \makeatletter
 \DeclareRobustCommand\onedot{\futurelet\@let@token\@onedot}
 \def\@onedot{\ifx\@let@token.\else.\null\fi\xspace}
 \def\eg{e.g\onedot} 
 \def\ie{i.e\onedot}

 \makeatother

\newcommand{\myparagraph}[1]{\vspace{5pt}\noindent{\bf #1}}
\usepackage[pagebackref=true,breaklinks=true,letterpaper=true,colorlinks,bookmarks=false]{hyperref}

\cvprfinalcopy %

\def\cvprPaperID{2708} %

\ifcvprfinal\fi
\begin{document}

\title{Multimodal Explanations: Justifying Decisions and Pointing to the Evidence}

 \newcommand{\authSpace}{&}%
 \author{
 Dong Huk Park$^{1}$, Lisa Anne Hendricks$^{1}$, Zeynep Akata$^{2,3}$, Anna Rohrbach$^{1,3}$, \\
 Bernt Schiele$^{3}$, Trevor Darrell$^{1}$, and Marcus Rohrbach$^{4}$ \vspace{2mm} \\
  $^{1}$EECS, UC Berkeley,  $^2$University of Amsterdam,
 $^{3}$MPI for Informatics, $^{4}$Facebook AI Research \\
}

\newcommand{\toAdd}[1]{\textcolor{red}{\textbf{Todo #1}}}
\maketitle
\begin{abstract}
Deep models that are both effective and explainable are desirable in many settings; prior explainable models have been unimodal, offering either image-based visualization of  attention weights or text-based generation of post-hoc  justifications. We propose a multimodal approach to explanation, and argue that the two modalities provide complementary explanatory strengths. We collect two new datasets to define and evaluate this task, and propose a novel model which can provide joint textual rationale generation and attention visualization. Our datasets define visual and textual justifications of a classification decision for activity recognition tasks (ACT-X) and for visual question answering tasks (VQA-X). We quantitatively show that training with the textual explanations not only yields better textual justification models, but also better localizes the evidence that supports the decision. We also qualitatively show cases where visual explanation is more insightful than textual explanation, and vice versa, supporting our thesis that multimodal explanation models offer significant benefits over unimodal approaches. %

\end{abstract}

\section{Introduction}
\label{sec:intro}
Explaining decisions is an integral part of human communication, understanding, and learning, and humans naturally provide both deictic (pointing) and textual modalities in a typical explanation. We aim to build deep learning models that also are able to explain their decisions with similar fluency in both visual and textual modalities.  
Previous machine learning methods for explanation were able to provide a text-only explanation conditioned on an image in context of a task, or were able to visualize active intermediate units in a deep network performing a task, but were unable to provide explanatory text grounded in an image. 

We propose a new model which can jointly generate visual and textual explanations, using an attention mask to localize salient regions when generating textual rationales. 
We argue that to train effective models, measure the quality of the generated explanations, compare with other methods, and understand when methods will generalize, it is important to have access to ground truth human explanations. Unfortunately, there is a dearth of datasets which include examples of how humans justify specific decisions.
Thus, we collect two new datasets, ACT-X and VQA-X, which allow us to train and evaluate our novel model, which we call the Pointing and Justification Explanation (PJ-X) model. 
PJ-X is explicitly multimodal: it incorporates an explanatory attention step, which allows our model to both visually point to the evidence and justify a model decision with text. 

\begin{figure}[t]
\small 
\includegraphics[width=\linewidth]{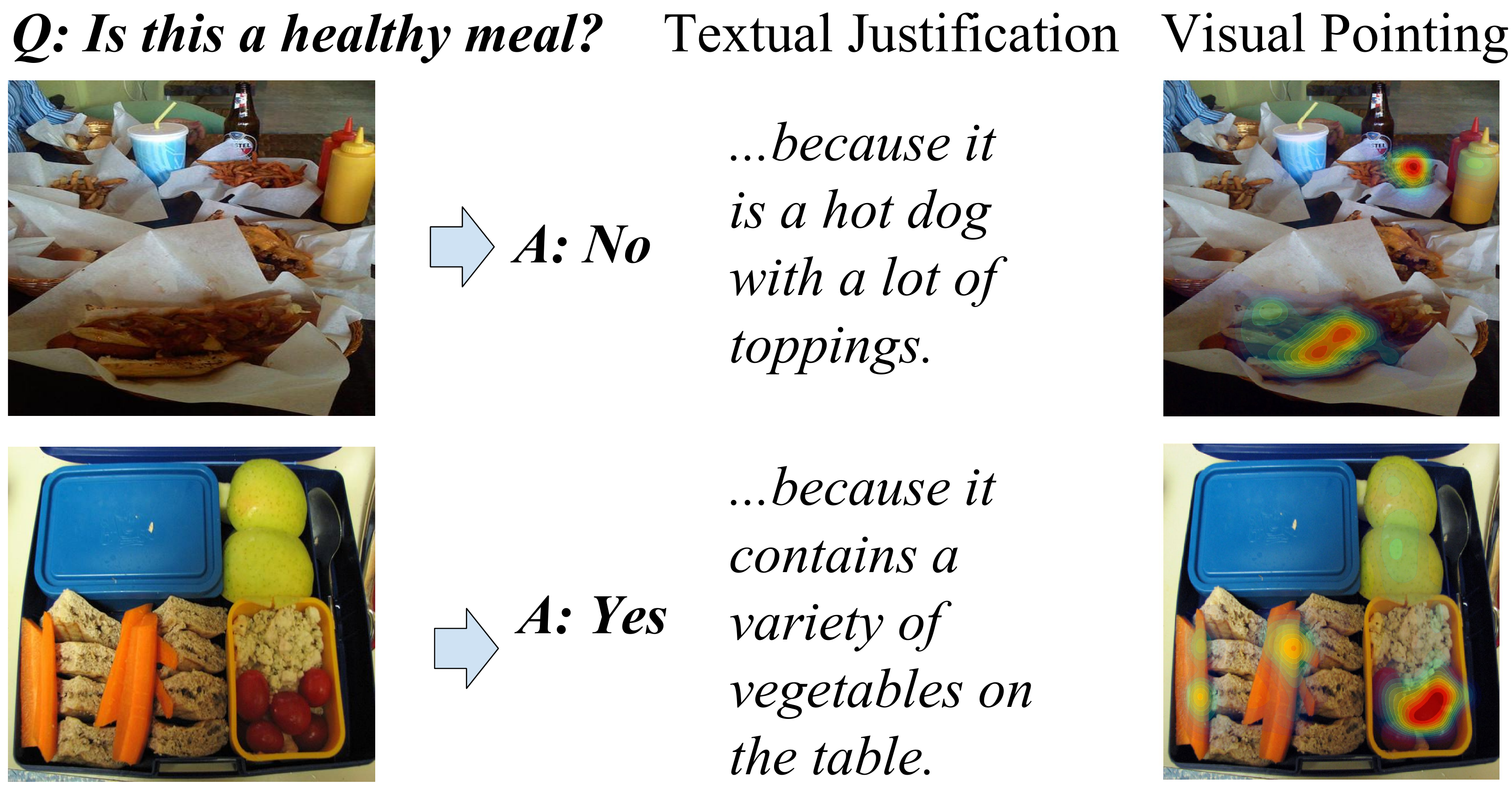}
\caption{For a given question and an image, our Pointing and Justification Explanation (PJ-X) model predicts the answer and \textit{multimodal} explanations which both point to the visual evidence for a decision and provide textual justifications. We show that considering multimodal explanations results in better visual and textual explanations.} 
\vspace{-0.3cm}
\label{fig:teaser}
\end{figure}

To illustrate the utility of multimodal explanations, consider \autoref{fig:teaser}. In both examples, the question ``Is this a healthy meal?'' is asked, and the PJ-X model correctly answers either ``no'' or ``yes'' depending on the visual input.
To justify why the image is not healthy, the generated textual justification mentions the kinds of unhealthy food in the image (``hot dog'' and ``toppings'').
In addition to mentioning the unhealthy food, our model is able to \textit{point} to the hot dog in the image.
Likewise, to justify why the image on the right is healthy, the textual explanation model mentions ``vegetables''.
Note that PJ-X model then points to the vegetables, which are mentioned in the textual explanation, but not other items in the image, such as the bread.

We propose VQA and activity recognition as testbeds for studying explanations because they are challenging and important visual tasks which have interesting properties for explanation.
VQA is a widely studied multimodal task that requires visual and textual understanding as well as commonsense knowledge. 
The newly collected VQA v2 dataset \cite{goyal17cvpr} includes complementary pairs of questions and answers. %
Complementary VQA pairs ask the same question of two semantically similar images which have different answers. As the two images are semantically similar, VQA models must employ finegrained reasoning to answer the question correctly. Not only is this an interesting and useful setting for measuring overall VQA performance, but it is also interesting when studying explanations. By comparing explanations from complementary pairs, we can more easily determine whether our explanations focus on the important factors for making a decision.

Additionally, we collect annotations for activity recognition using the MPII Human Pose (MHP) dataset \cite{APGS14}. Activity recognition in still images relies on a variety of cues, such as pose, global context, and the interaction between humans and objects. Though a recognition model can potentially classify an activity correctly, it is not capable of indicating which factors influence the decision process.
Furthermore, classifying specific activities requires understanding finegrained differences (e.g., ``road biking'' and ``mountain biking'' include similar objects like ``bike'' and ``helmet,'' but road biking occurs on a road whereas mountain biking occurs on a mountain path).
Such finegrained differences are interesting yet difficult to capture when explaining neural network decisions.

In sum, we present ACT-X and VQA-X, two novel datasets of human annotated multimodal explanations (visual and textual) for activity recognition and visual question answering. These datasets allow us to train the Pointing and Justification (PJ-X) model which goes beyond current visual explanation systems by producing \textit{multimodal} explanations, justifying the predicted answer post-hoc by visual pointing and textual justification.  
Our datasets also allow to effectively evaluate explanation models, and we show that the PJ-X model outperforms strong baselines, and, importantly, that by generating multimodal explanations, we outperform models which only produce visual or textual explanations.
We will release our model architecture, learned weights, and datasets upon acceptance of this paper.
\section{Related Work}
\label{sec:related}

\myparagraph{Explanations.}  
Early textual explanation models span a variety of applications (e.g., medical \cite{shortliffe1975model} and feedback for teaching programs \cite{lane2005explainable, van2004explainable, core2006building}) and are generally template based.  More recently, \cite{hendricks16eccv} developed a deep network to generate natural language justifications of a fine-grained object classifier. 
However, unlike our model, it does not provide multimodal explanations.
Furthermore, \cite{hendricks16eccv} could not train on reference human explanations as no such datasets existed. 
We provide two datasets with reference textual explanations to enable more research in the direction of textual explanation generation.

A variety of work has proposed methods to visually explain decisions.  Some methods find discriminative visual patches  \cite{doersch2012makes,berg2013you} whereas others aim to understand intermediate features which are important for end decisions~\cite{zeiler2014visualizing,escorcia2015relationship,zhou2014object} \eg what does a certain neuron represent.
Our model PJ-X points to visual evidence via an attention mechanism which is an intuitive way to convey knowledge about what is important to the network without requiring domain knowledge.
Unlike prior work, PJ-X generates \emph{multimodal explanations} in the form of explanatory sentences and attention maps pointing to the visual evidence.

Prior work has investigated how well generated visual explanations align with human gaze~\cite{DBLP:journals/corr/DasAZPB16}.
However, when answering a question, humans do not always look at image regions which are necessary to explain a decision. 
For example, given the question ``What is the name of the restaurant?'', human gaze might capture other buildings before settling on the restaurant.
In contrast, when we collect our annotations, we allow annotators to view the entire image and ask them to point to the most relevant visual evidence for making a decision.
Furthermore, our visual explanations are collected in conjunction with textual explanations to build and evaluate multimodal explanation models.

\begin{figure}[t]
\center
  \includegraphics[width=0.9\linewidth]{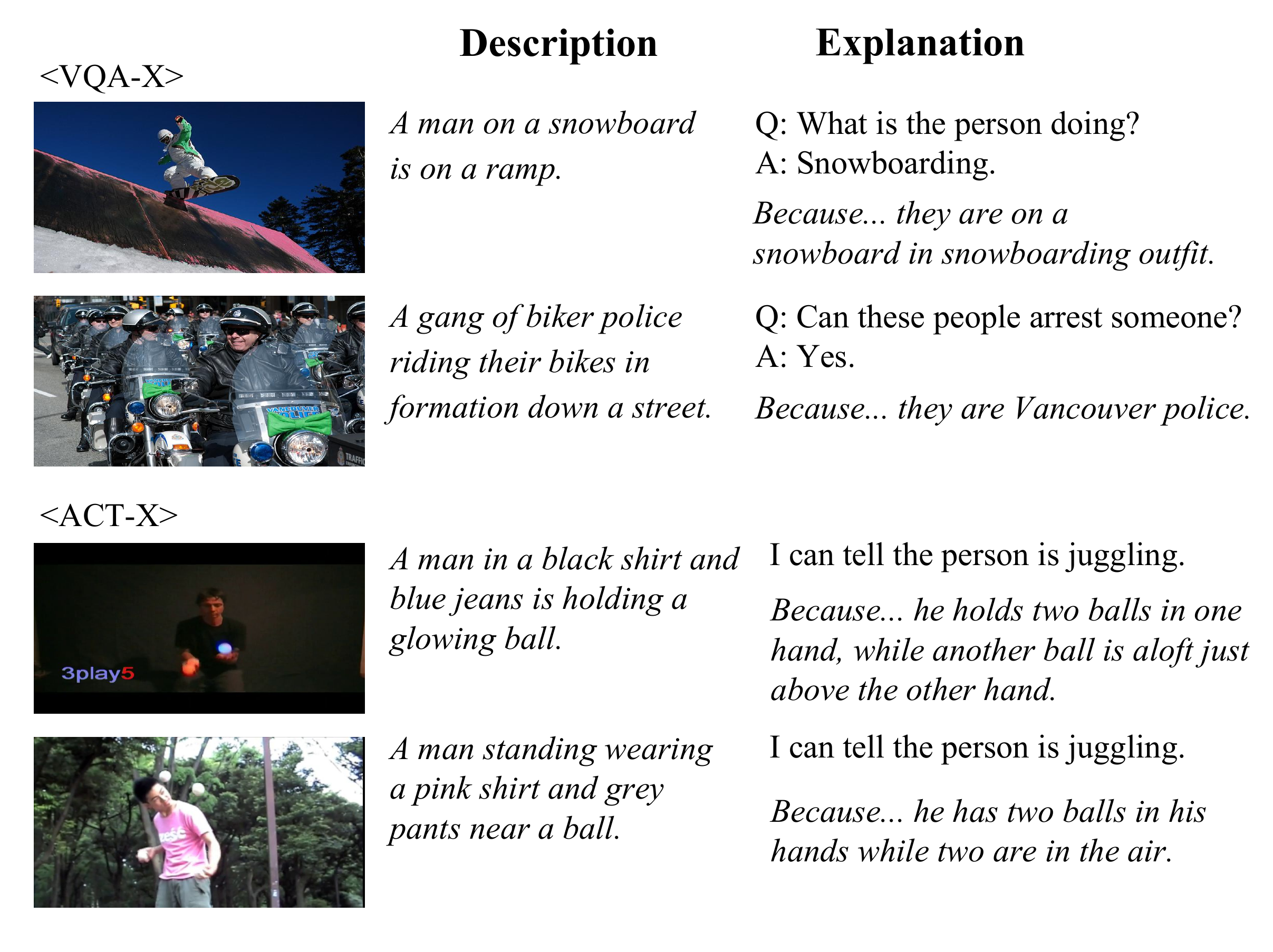}
\caption{In comparison to descriptions, our VQA-X explanations focus on the evidence that pertains to the {\em question and answer} instead of generally describing the scene. For ACT-X, our explanations are task specific whereas descriptions are more generic.}
\label{fig:X_dataset_examples}
\vspace{-0.3cm}
\end{figure}

{
\renewcommand{\arraystretch}{1.1}

\begin{table*}[t]
\small
\center
  \begin{tabular}{@{ }c@{\ } | l@{\ \ \ \ \ \ }c@{\ \ }c@{\ \ }c@{\ \ }c@{\ \ }c@{\ }c@{\ }c@{\ \ }c@{\ \ }c@{\ }} 
    \textbf{Dataset} & \textbf{Split} & \#Imgs & \#Q/A Pairs & \#Unique Q. & \#Unique A. & \#Expl.&(Avg. \#w) & Expl.Vocab Size & \#Comple. Pairs &\#Visual Ann.\\     \hline
   \multirow{4}{*}{VQA-X} & Train & $24876$ & $29459$ & $12942$ & $1147$ & $31536$ & ($8.56$) & $12412$ & $6050$ & --  \\
   & Val & $1431$ & $1459$ & $813$ & $246$ & $4377$ & ($8.89$) & $4325$ & $240$ & $3000$ \\
   & Test & $1921$ & $1968$ & $898$ & $272$ & $5904$ & ($8.94$) & $4861$ & $510$ & $3000$ \\
\cline{2-11}
   & Total & $28180$ & $32886$ & $13921$ & $1236$ & $41817$ & ($8.64$) & $14106$ & $6800$ & $6000$ \\
\hline\hline
  \multirow{4}{*}{ACT-X} & Train & $12607$ & -- & -- & $397$ & $37821$ & ($13.96$) & $12377$ & -- & -- \\
  & Val & $1802$ & -- & -- & $295$ & $5406$ & ($13.91$) & $4802$ & -- & $3000$ \\
  & Test & $3621$ & -- & -- & $379$ & $10863$ & ($13.96$) & $6856$ & -- & $3000$ \\
\cline{2-11}
  & Total & $18030$ & -- & -- & $397$ & $54090$ & ($13.95$) & $14588$ & -- &  $6000$ \\
\bottomrule
  \end{tabular}
  \vspace{-0.1cm}
\caption{Dataset statistics for VQA-X (top) and ACT-X (bottom).
{\small Unique Q. = Unique questions, Unique A. = Unique answers, Expl. = Explanations, Avg. \#w = Average number of words, Comple. Pairs = Complementary pairs, Visual Ann. = Visual annotations.}
}
\label{tab:datasets}
\end{table*}
}

\myparagraph{Visual Question Answering and Attention.}
Initial approaches to VQA used full-frame representations \cite{malinowski15iccv}, but most recent approaches use some form of spatial attention~\cite{yang2015stacked,xu2015ask,zhu16cvpr,chen2015abc,xiong16dynamic,shih2015look,fukui16emnlp,DBLP:journals/corr/KimOKHZ16}. We base our method on \cite{fukui16emnlp}, the winner of VQA 2016 challenge, %
however we use an element-wise product as opposed to compact bilinear pooling.
\cite{DBLP:journals/corr/KimOKHZ16} has explored the element-wise product for VQA just as we do in our method, however~\cite{DBLP:journals/corr/KimOKHZ16} improves performance by applying hyperbolic tangent (TanH) after the multimodal pooling whereas we improve by applying signed square-root and L2 normalization.

\myparagraph{Activity Recognition.}
Recent work on activity recognition in still images relies on a variety of cues, such as pose and global context \cite{gkioxari2015contextual,mallya16eccv,pishchulin14gcpr}.
Specifically, \cite{gkioxari2015contextual} considers additional image regions and \cite{mallya16eccv} considers a global image feature in addition to the region where an activity occurs.
Generally, works on the MPII Human Activities dataset provide the ground truth location of a human at test time \cite{gkioxari2015contextual}.
In contrast, we consider a more realistic scenario and do not make any assumptions about where the activity occurs at test time.
Our model relies on attention to focus on important parts of an image for classification and explanation.

{

\section{Multimodal Explanations}
\label{sec:datasets}
We propose multimodal explanation tasks with visual and textual components, defined on both visual question answering and activity recognition testbeds.
To train and evaluate models for this task we collect two multimodal explanation datasets: Visual Question Answering Explanation (VQA-X) and Activity Explanation (ACT-X) (see ~\autoref{tab:datasets} for a summary). For each dataset we collect textual and visual explanations from human annotators.

\myparagraph{VQA Explanation Dataset (VQA-X).} The Visual Question Answering (VQA) dataset~\cite{antol2015vqa} contains open-ended questions about images which require understanding vision, language, and commonsense knowledge to answer. VQA consists of approximately $200$K MSCOCO images~\cite{lin2014microsoft}, with $3$ questions per image and $10$ answers per question. %

\begin{figure*}
\center
\vspace{-0.3cm}
\subfigure[VQA-X]{\label{fig:vqa_seg_gt}\includegraphics[width=70mm, height=55mm]{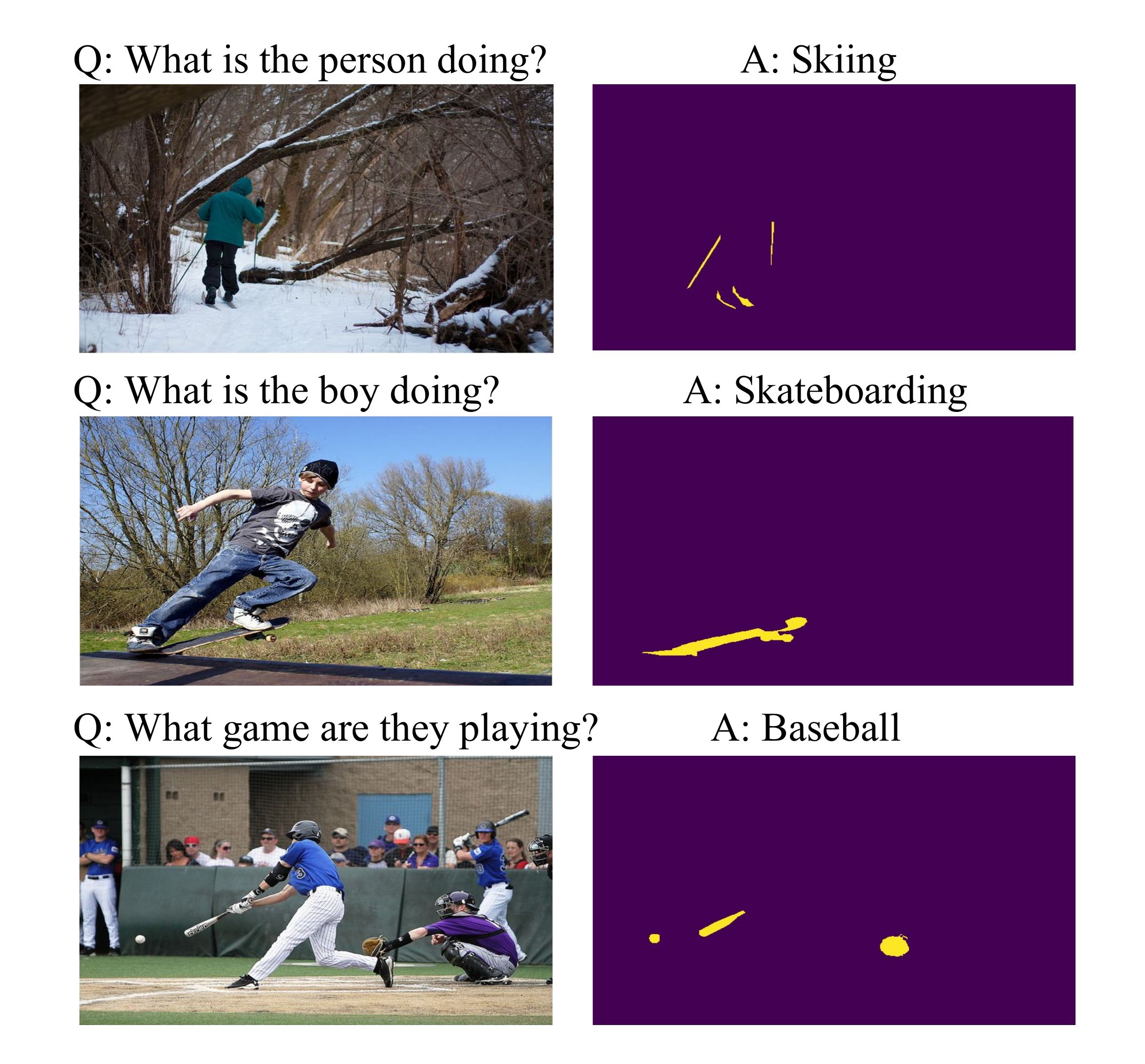}}
\subfigure[ACT-X]{\label{fig:act_seg_gt}\includegraphics[width=70mm, height=55mm]{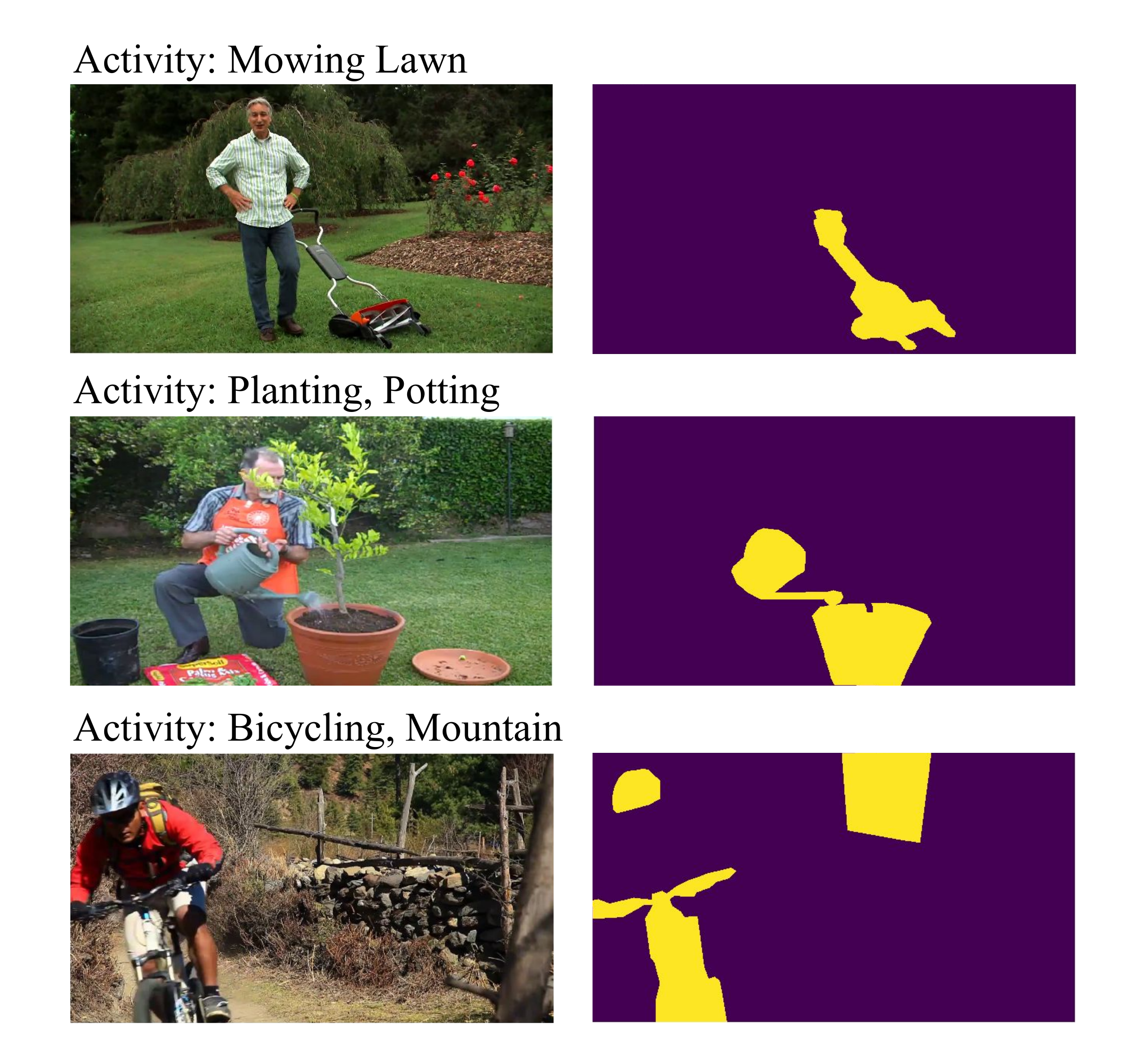}}
\vspace{-0.3cm}
\caption{Human annotated visual explanations. On the left: example annotations collected on VQA-X dataset. On the right: Example annotations collected on ACT-X dataset. The visual evidence that justifies the answer is segmented in yellow.}
\label{fig:GroundTruthMap}
\end{figure*}

\begin{figure}[t]
\center
  \vspace{-0.7cm}
  \includegraphics[width=0.8\linewidth]{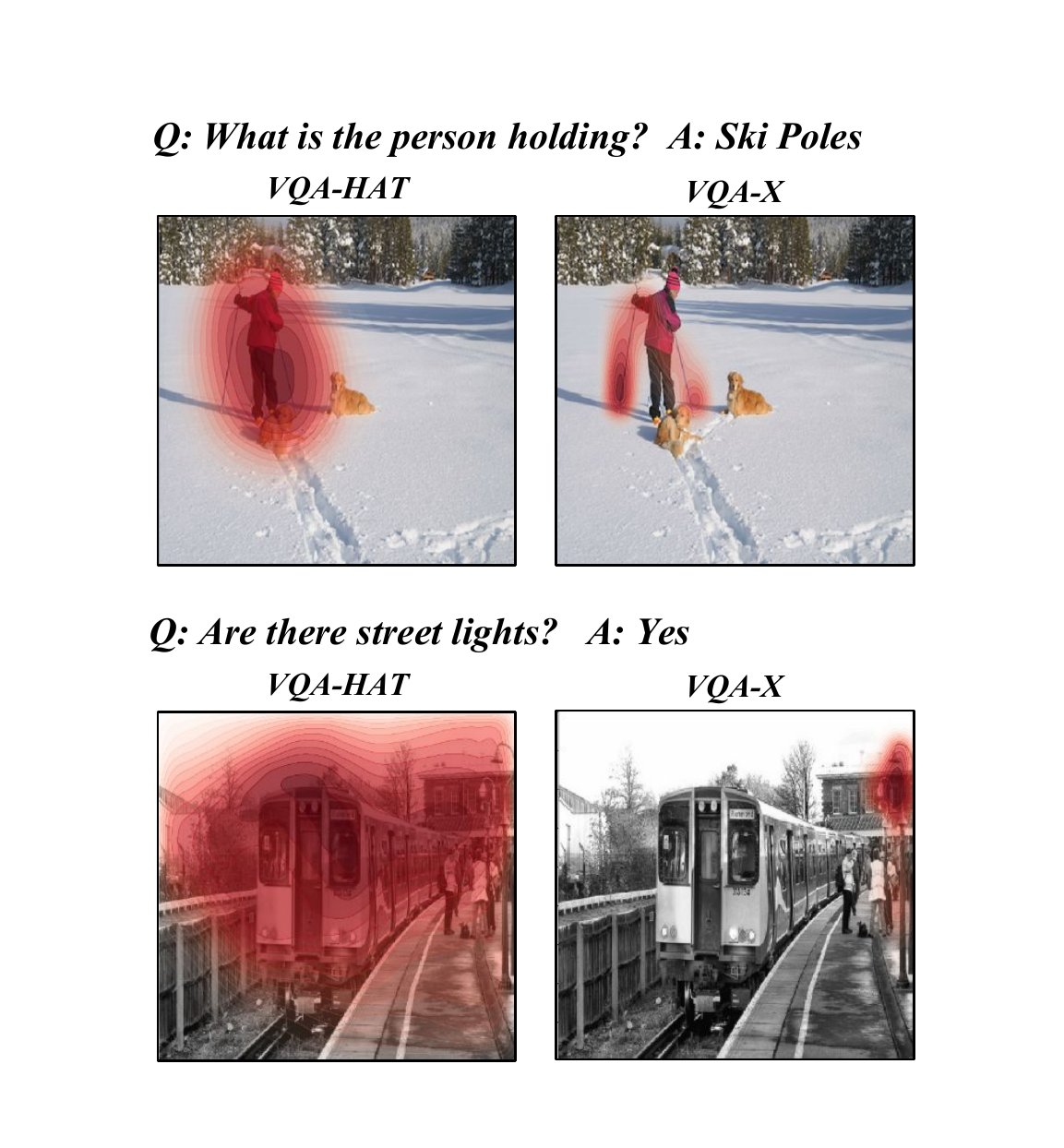}
  \vspace{-0.6cm}
\caption{Human visual annotations from VQA-HAT and VQA-X. We aggregate all the annotations in each image and normalize them to create a probability distribution. The distribution is then visualized over the image as a heatmap.}
\label{fig:vqa_hat_vs_ours}
\end{figure}

Many questions in VQA are of the sort: ``What is the color of the banana?''. It is difficult for humans to explain answers to such questions because it requires explaining a fundamental visual property: color.
Thus, we aim to provide textual explanations for questions that go beyond such trivial cases. 
To do this, we consider the annotations collected in \cite{DBLP:journals/corr/ZitnickAAMBP16} which say how old a human must be to answer a question. We find that questions which require humans to be of age 9 or higher are generally interesting to explain. 

Additionally, we consider complementary pairs from the VQA v2 dataset~\cite{goyal17cvpr}.
Complementary pairs consist of a question and two similar images which give two different answers.
Complementary pairs are particularly interesting for the explanation task because they allow us to understand whether explanations name the correct evidence based on image content, or whether they just memorize which content to consider based off specific question types.

We collect a single textual explanation for QA pairs in the training set and three textual explanations for test/val QA pairs. Some examples can be seen \autoref{fig:X_dataset_examples}.

\myparagraph{Action Explanation Dataset (ACT-X).} The MPII Human Pose (MHP) dataset~\cite{APGS14} contains $25$K images extracted from Youtube videos. From the MHP dataset, we select all images that pertain to $397$ activities, resulting in $18,030$ images total.
For each image we collect three explanations. During data annotation, we ask the annotators to complete the sentence ``I can tell the person is doing (action) because..'' where the action is the ground truth activity label. We also ask them to use at least 10 words and avoid mentioning the activity class in the sentence. %
MHP dataset also comes with sentence descriptions provided by~\cite{RAMTSL16}. See \autoref{fig:X_dataset_examples} for examples of descriptions and explanations.

\myparagraph{Ground truth for pointing.}
In addition to textual justifications, we collect visual explanations from humans for both VQA-X and ACT-X datasets in order to evaluate how well the attention of our model corresponds to where humans think the evidence for the answer is. 
Human-annotated visual explanations are collected via Amazon Mechanical Turk where we use the segmentation UI interface from the OpenSurfaces Project \cite{bell13opensurfaces}. 
Annotators are provided with an image and an answer (question and answer pair for VQA-X, class label for ACT-X). 
They are asked to segment objects and/or regions that most prominently justify the answer. 
For each dataset we randomly sample images from the test split, and for each image we collect 3 annotations. 
Some examples can be seen in \autoref{fig:GroundTruthMap}.

\begin{figure*}[t]
\includegraphics[width=\linewidth]{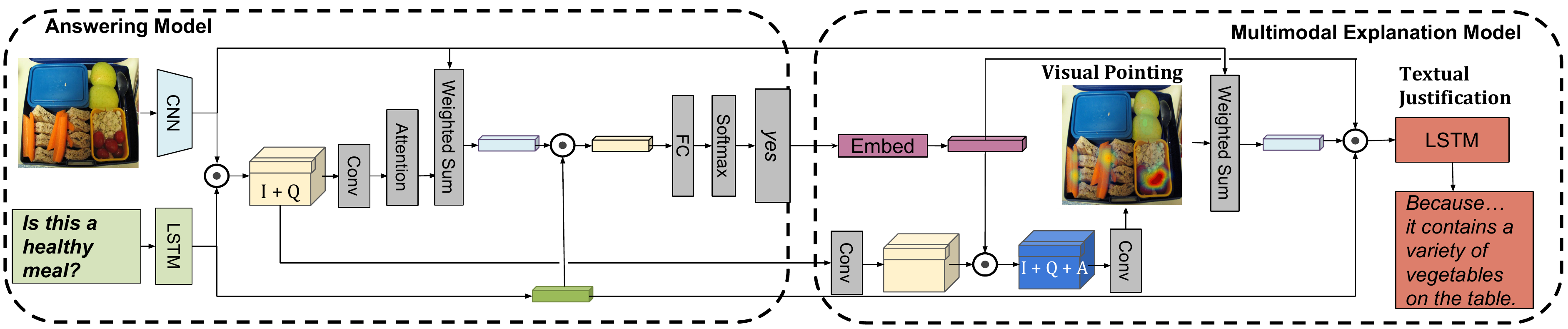}
\caption{Our Pointing and Justification (PJ-X)
architecture generates a multimodal explanation which includes textual justification (``it contains a variety of vegetables on the table'') and pointing to the visual evidence.}
\label{fig:attentive_explanation}
\end{figure*}

\myparagraph{Comparing with VQA-HAT.}
A thorough comparison between our dataset and VQA-HAT dataset from~\cite{DBLP:journals/corr/DasAZPB16} is currently not viable because the two datasets have different splits and the overlap is small. However, we present qualitative comparison in \autoref{fig:vqa_hat_vs_ours}. In the first row, our VQA-X annotation has a finer granularity since it segments out the objects in interest more accurately than the VQA-HAT annotation. In the second row, our annotation contains less extraneous information than the VQA-HAT annotation. Since the VQA-HAT annotations are collected by having humans ``unblur'' the images, they are more likely to introduce noise when irrelevant regions are uncovered. 
\section{Pointing and Justification Model (PJ-X)}
\label{sec:method}

The goal of our work is to implement a multimodal explanation system that justifies a decision with natural language and points to the evidence.
We deliberately design our Pointing and Justification Model (PJ-X) to allow training these two tasks.
Specifically we want to rely on natural language justifications and the classification labels as the only supervision. %
We design the model to learn how to point in a latent way. For the pointing we rely on an attention mechanism \cite{bahdanau2014neural} which allows the model to focus on a spatial subset of the visual representation. 

We first predict the answer given an image and question using the answering model. Then given the answer, question, and image, we generate visual and textual explanations with the multimodal explanation model. 
An overview of our model is presented in \autoref{fig:attentive_explanation}.

\myparagraph{Answering model.} In visual question answering the goal is to predict an answer given a question and an image. For activity recognition we do not have an explicit question. %
Thus, we ignore the question which is equivalent to setting the question representation to $f^Q(Q)=1$, a vector of ones. %
We base our answering model on the overall architecture from the MCB model~\cite{fukui16emnlp}, but %
replace the MCB unit 
with a simpler element-wise multiplication $\odot$ to pool multimodal features. 
We found that this leads to similar performance, but much faster training (see supplemental material). 

In detail, we extract spatial image features $f^{I}(I,n,m)$ from the last convolutional layer of ResNet-152 followed by $1\times1$ convolutions ($\bar{f}^{I}$) giving a $2048\times N\times M$ spatial image feature. We encode the question $Q$ with a 2-layer $LSTM$, which we refer to as $f^Q(Q)$. We combine this and the spatial image feature using element-wise multiplication followed by signed square-root, L2 normalization, and Dropout, and two more layers of $1\times1$ convolutions with ReLU in between.
This process gives us a $N\times M$ attention map $\bar{\alpha}_{n,m}$. We apply softmax to produce a normalized soft attention map.

The attention map is then used to take the weighted sum over the image features and this representation is once again combined with the LSTM feature to predict the answer $\hat{y}$ as a classification problem over all answers $Y$. We provide an extended formalized version in the supplemental.

\paragraph{Multimodal explanation model.}
We argue that to generate multimodal explanation, we should condition it on question, answer, and image. For instance, to be able to explain ``Because they are Vancouver police'' in~\autoref{fig:X_dataset_examples}, the model needs to see the question, \ie ``Can these people arrest someone?'', the answer, \ie ``Yes'' and the image, \ie the ``Vancouver police'' banner on the motorcycles. 

We model this by pooling image, question, and answer representations to generate  attention map, our \emph{Visual Pointing}. The \emph{Visual Pointing} is further used to create attention features that guide the generation of our \emph{Textual Justification}.

More specifically, the answer predictions are embedded in a $d$-dimensional space followed by $\tanh$ non-linearity and a fully connected layer:
$f^{yEmbed}(\hat{y}) =  W_6(tanh(W_5 \hat{y} + b_5))+b_6$.
To allow the model to learn how to attend to relevant spatial location based on the answer, image, and question, we combine this answer feature with Question-Image embedding $\bar{f}^{IQ}(I,Q)$ from the answering model. %
Applying $1\times1$ convolutions, element-wise multiplication followed by signed square-root, L2 normalization, and Dropout, results in a multimodal feature. %
\begin{align}
\bar{f}^{IQA}(I,n,m,Q,\hat{y}) = & (W_7 \bar{f}^{IQ}(I,Q,n,m) + b_7) \\
&\odot f^{yEmbed}(\hat{y}))\\
f^{IQA}(I,Q,\hat{y}) =& L2(\textit{signed\_sqrt}(\bar{f}^{IQA}(I,Q,\hat{y})))
\end{align}
with Relu $\rho(x) = max(x,0)$.
Next we predict a $N\times M$ attention map $\bar{\alpha}_{n,m}$ 
and apply softmax to produce a normalized soft attention map, our \emph{Visual Pointing } {\emph{$\alpha_{n,m}^{pointX}$}}, which aims to point at the evidence of the generated explanation:
\begin{align}
\label{eq:attention}
\bar{\alpha}_{n,m} = &f^{pointX}(I,n,m,Q,\hat{y})\\
= & W_{9}\rho(W_{8} f^{IQA}(I,Q,\hat{y}) + b_{8}) + b_{9}\\
\alpha_{n,m}^{pointX} = &\frac{\exp(\bar{\alpha}_{n,m})}{\sum_{i=1}^N\sum_{j=1}^M{\exp(\bar{\alpha}_{n,m})}}
\end{align}

Using {\emph{$\alpha_{n,m}^{pointX}$}}, we compute the attended visual representation, and merge it with the LSTM feature that encodes the question and the embedding feature that encodes the answer:
\begin{align}
\
f^{X}(I,Q,\hat{y}) = & (W_{10} \sum_{x=1}^N\sum_{y=1}^M{\alpha_{n,m}^{pointX}f^{I}(I,n,m) } + b_{10})\\
& \odot (W_{11} f^Q(Q) + b_{11})  \odot f^{yEmbed}(\hat{y})
\end{align}

This combined feature is then fed into an LSTM decoder to generate our Textual Justifications  that are conditioned on image, question, and answer.

\emph{Textual Justifications}  are a sequence of words {\boldmath $[w_1,w_2,\ldots]$} and our model predicts one word $w_{t}$ at each time step $t$  conditioned on the previous word and the hidden state of the LSTM:
\begin{align}
h_{t} = f^{LSTM}(f^{X}(I,Q,\hat{y}),w_{t-1},h_{t-1})\\
w_{t} = f^{pred}( h_{t}) = Softmax(W_{pred} h_{t} + b_{pred}  )
\end{align}

\section{Experiments}
\label{sec:exp}

In this section, after detailing the experimental setup, 
we present quantitative results on ablations done for textual justification and visual pointing tasks, and discuss their implications. Additionally, we provide and analyze qualitative results for both tasks.

\subsection{Experimental Setup}
\label{sec:experimentalsetup}
Here, we detail our experimental setup in terms of model training, hyperparameter settings, and evaluation metrics.

\myparagraph{Model training and hyperparameters.}
For VQA, the answering model of PJ-X is pre-trained on the VQA v2 training set \cite{goyal17cvpr}. 
We then freeze or finetune the weights of the answering model when training the multimodal explanation model on textual annotations as the VQA-X dataset is significantly smaller than the original VQA training set. For activity recognition, answering and explanation components of PJ-X are trained jointly. 
The spatial feature size of PJ-X is $N=M=14$. 
For VQA, we limit the answer space to the $3000$ most frequently occurring answers on the training set (\ie $|Y|=3000$) whereas for activity recognition, $|Y|=397$. We set the answer embedding size as $d=300$ for both tasks.

\myparagraph{Evaluation metrics.} We evaluate our textual justifications w.r.t BLEU-4~\cite{papineni2002bleu}, METEOR~\cite{banerjee2005meteor}, ROUGE~\cite{L04}, CIDEr~\cite{vedantam2015cider} and SPICE~\cite{spice2016} metrics, which measure the degree of similarity between generated and ground truth sentences.
We also include human evaluation since automatic metrics do not always reflect human preference. We randomly choose 1000 data points each from the test splits of VQA-X and ACT-X datasets, where the model predicts the correct answer, and then for each data point ask 3 human subjects to judge whether a generated explanation is better than, worse than, or equivalent to the ground truth explanation (we note that human judges do not know what explanation is ground truth and the order of sentences is randomized).
We report the percentage of generated explanations which are equivalent to or better than ground truth human explanations, when at least 2 out of 3 human judges agree.

For visual pointing task, we use Earth Mover's Distance (EMD)~\cite{RubnerTG98iccv} which measures the distance between two probability distributions over a region. We use the code from \cite{pele2009} to compute EMD.
We also report on Rank Correlation which was used in~\cite{DBLP:journals/corr/DasAZPB16}.
For computing Rank Correlation, we follow~\cite{DBLP:journals/corr/DasAZPB16} where we scale the generated attention map and the human ground-truth annotations from the VQA-X/ACT-X/VQA-HAT datasets to $14 \times 14$, rank the pixel values, and then compute correlation between these two ranked lists.

{
\setlength{\tabcolsep}{3pt}
\renewcommand{\arraystretch}{1.2}
\begin{table*}[tb]
\begin{center}
\small
\begin{tabular}{l l l l|rrrrr c|rrrrr c}
\toprule
&GT-ans&Train-&Att.&\multicolumn{6}{c|}{VQA-X} & \multicolumn{6}{c}{ACT-X} \\
&Condi-&ing &  for & \multicolumn{5}{c}{Automatic evaluation} & \multicolumn{1}{c|}{Human} &\multicolumn{5}{c}{Automatic evaluation} & \multicolumn{1}{c}{Human}\\
Approach &  tioning & Data & Expl. & \multicolumn{1}{c}{B} & \multicolumn{1}{c}{M} & \multicolumn{1}{c}{R} & \multicolumn{1}{c}{C} &\multicolumn{1}{c}{S} &\multicolumn{1}{c|}{eval} & \multicolumn{1}{c}{B} & \multicolumn{1}{c}{M} & \multicolumn{1}{c}{R} & \multicolumn{1}{c}{C} &\multicolumn{1}{c}{S} &\multicolumn{1}{c}{eval}  \\
\midrule
\cite{hendricks16eccv} & Yes & Desc. & No &\multicolumn{1}{c}{--}&\multicolumn{1}{c}{--}&\multicolumn{1}{c}{--}&\multicolumn{1}{c}{--}&\multicolumn{1}{c}{--}&--&12.9 & 15.9  & 39.0 & 12.4 & 12.0 &17.4 \\ %
Ours on Descriptions & Yes & Desc. & Yes & 6.1 & 12.8 & 26.4  & 36.2 & 12.1 &34.5&6.9  & 12.9 & 28.3 & 20.3 & 7.3& 22.9 \\ %
Ours w/o Attention & Yes & Expl. & No & 18.0 & 17.6 & 42.4 & 66.3 & 14.3 &40.1& 16.9 &17.0 &42.0&33.3 &10.6 &21.4 \\
Ours & Yes & Expl. & Yes  & \textbf{19.8} & \textbf{18.6} & \textbf{44.0} & \textbf{73.4} & \textbf{15.4} &\textbf{45.1}& \textbf{24.5} &\textbf{21.5} &\textbf{46.9} &\textbf{58.7} &\textbf{16.0} &\textbf{38.2}\\
\midrule
Ours on Descriptions & No& Desc. & Yes & 5.9 & 12.6  & 26.3  & 35.2 & 11.9 &-- & 5.2  & 11.0 & 26.5 & 10.4 & 4.6& -- \\
Ours w/o Attention & No&  Expl. & No & 18.0 & 17.3 & 42.1 & 63.6 & 13.8 &-- & 11.9 & 13.6 &37.9&16.9 &5.7 & -- \\ %
Ours & No & Expl. & Yes & \textbf{19.5} & \textbf{18.2} & \textbf{43.4} & \textbf{71.3} & \textbf{15.1} & -- & \textbf{15.3} &\textbf{15.6} &\textbf{40.0} &\textbf{22.0} &\textbf{7.2} &  -- \\ 
\bottomrule
\end{tabular}
\vspace{-2mm}
\caption{Evaluation of Textual Justifications. Evaluated automatic metrics: BLEU-4 (B), METEOR (M), ROUGE (R), CIDEr (C) and SPICE (S). Reference sentence for human and automatic evaluation is always an explanation. All in \%.  Our proposed model compares favorably to baselines.}
\vspace{-4mm}
\label{tbl:generation_automatic}
\end{center}
\end{table*}
}

\subsection{Textual Justification}
\label{sec:res:textual}
We ablate PJ-X and compare with related approaches on our VQA-X and ACT-X datasets through automatic and human evaluations for the generated explanations.

\myparagraph{Details on compared models.} 
We compare with the state-of-the-art~\cite{hendricks16eccv} using publicly available code. 
For fair comparison, we use ResNet features extracted from the entire image when training \cite{hendricks16eccv}.
The generated sentences from \cite{hendricks16eccv} are conditioned on both the image and the class label. \cite{hendricks16eccv} uses discriminative loss, which enforces the generated sentence to contain class-specific information, to back-propagate policy gradients when training the language generator, and thus involves training a separate sentence classifier to generate rewards. Our model does not use discriminative loss/policy gradients and does not require defining a reward.
Note that \cite{hendricks16eccv} is trained with descriptions.
Similarly, ''Ours on Descriptions'' is an ablation in which we train PJ-X on descriptions instead of explanations. 
''Ours w/o Attention'' is similar to~\cite{hendricks16eccv} in the sense that there is no attention mechanism involved when generating explanations, however, it does not use the discriminative loss and is trained on explanations instead of descriptions. 

\myparagraph{Descriptions vs. Explanations.} ``Ours'' significantly outperforms ``Ours with Descriptions'' by a large margin on both datasets which is expected as descriptions are insufficient for the task of generating explanations. 
Additionally, ``Ours'' compares favorably to \cite{hendricks16eccv} even in the case when ``Ours'' generates textual justifications conditioned on the prediction, not the ground-truth answer. These results demonstrate the limitation of training explanation systems with descriptions, and thus support the necessity of having datasets specifically curated for explanations. 
``Ours on Descriptions'' performs worse on certain metrics compared to~\cite{hendricks16eccv} which may be attributed to additional training signals generated from discriminative loss and policy gradients, but further investigation is left for future work. 

\myparagraph{Unimodal explanations vs. Multimodal explanations.} 
Including attention when generating textual justifications allows us to build a multimodal explanation model. Aside from the immediate benefit of providing visual rationale about a model's decision, learning to point at visual evidence helps generating better textual justifications.
As can be seen from~\autoref{tbl:generation_automatic}, ``Ours'' greatly improves textual justifications compared to ``Ours w/o Attention'' on both datasets, demonstrating the value of designing multimodal explanation systems.

\begin{table}[tb]
\setlength{\tabcolsep}{2pt}
\renewcommand{\arraystretch}{1.2}
\small
\centering
\begin{tabular}{l|rr|rrr}
\toprule
& \multicolumn{2}{c|}{Earth Mover's } & \multicolumn{3}{c}{ Rank Correlation }  \\
& \multicolumn{2}{c|}{(lower is better)} & \multicolumn{3}{c}{(higher is better) } \\
& VQA-X & ACT-X & VQA-X & ACT-X & VQA-HAT \\
\midrule
Random Point & 6.71 & 6.59 & +0.0017 & +0.0003 & -0.0001 \\
Uniform & 3.60 &  3.25 & +0.0003 & -0.0001 & -0.0007 \\
HieCoAtt-Q~\cite{DBLP:journals/corr/DasAZPB16} & -- & -- &-- & -- & +0.2640 \\
Answering Model  & 2.77 & 4.78 & +0.2211 & +0.0104 & +0.2234 \\
Ours  & \textbf{2.64} & \textbf{2.54} & \textbf{+0.3423} & \textbf{+0.3933} & \textbf{+0.3964} \\
\bottomrule
\end{tabular}

\caption{Evaluation of Visual Pointing Justifications.
  For rank correlation, all results have standard error $< 0.005$.
}
\label{tbl:attention_map_eval_wo}
\end{table}

\subsection{Visual Pointing}
\label{sec:res:pointing}
We compare the visual pointing performance of PJ-X to several baselines and report quantitative results with corresponding analysis.

\myparagraph{Details on compared models.}
We compare our model against the following baselines. \textit{Random Point} randomly attends to a single point in a $14\times 14$ grid. \textit{Uniform Map} generates attention map that is uniformly distributed over the $14\times 14$ grid. In addition to these baselines, we also compare PJ-X attention maps with those generated from state-of-the-art VQA systems such as \cite{DBLP:journals/corr/DasAZPB16}. %

\myparagraph{Improved localization with textual explanations.} We evaluate attention maps using the Earth Mover's Distance (lower is better) and Rank Correlation (higher is better) on VQA-X and ACT-X datasets in~\autoref{tbl:attention_map_eval_wo}. From ~\autoref{tbl:attention_map_eval_wo}, we observe that ``Ours'' not only outperforms baselines \textit{Random Point} and \textit{Uniform Map}, but also our answering model and \cite{DBLP:journals/corr/DasAZPB16} on both datasets and on both metrics. The attention maps generated from our answering model and \cite{DBLP:journals/corr/DasAZPB16} do not receive training signals from the textual annotations as they are only trained to predict the correct answer, whereas the attention maps generated from PJ-X multimodal explanation model are latently learned through supervision of textual annotations. The experiment results imply that learning to generate textual explanations helps improve visual pointing task, and further confirm the advantage of having a multimodal explanation system. 

\begin{figure}[t]
\center
  \includegraphics[width=.9\linewidth]{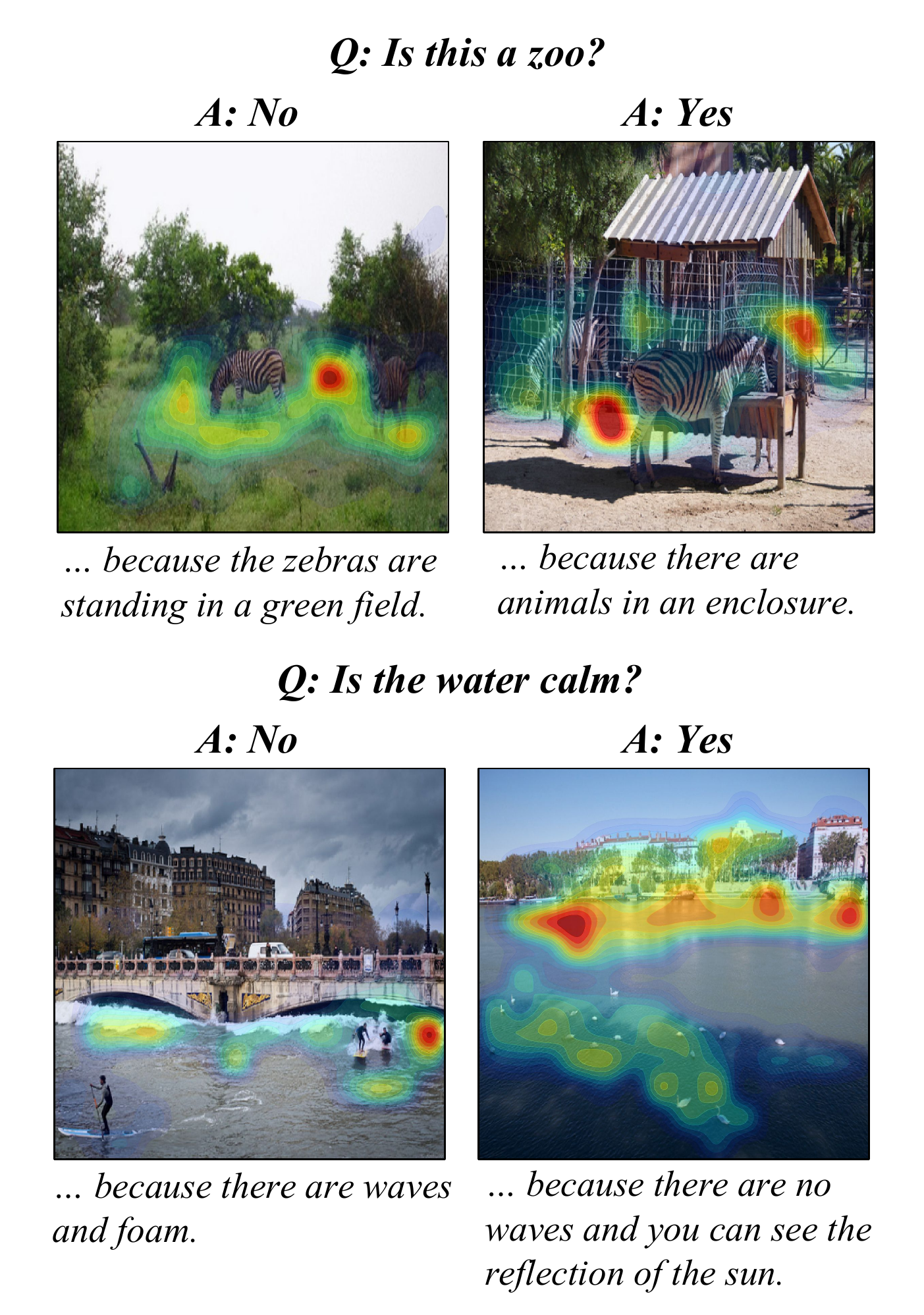}
\caption{VQA-X qualitative results: For each image the PJ-X model provides an answer and a justification, and points to the evidence for that justification. We show pairs of images from VQA v2 complementary pairs.}
\label{fig:VQAqualitative}
\end{figure}

\begin{figure}[t]
\center
  \includegraphics[width=.8\linewidth]{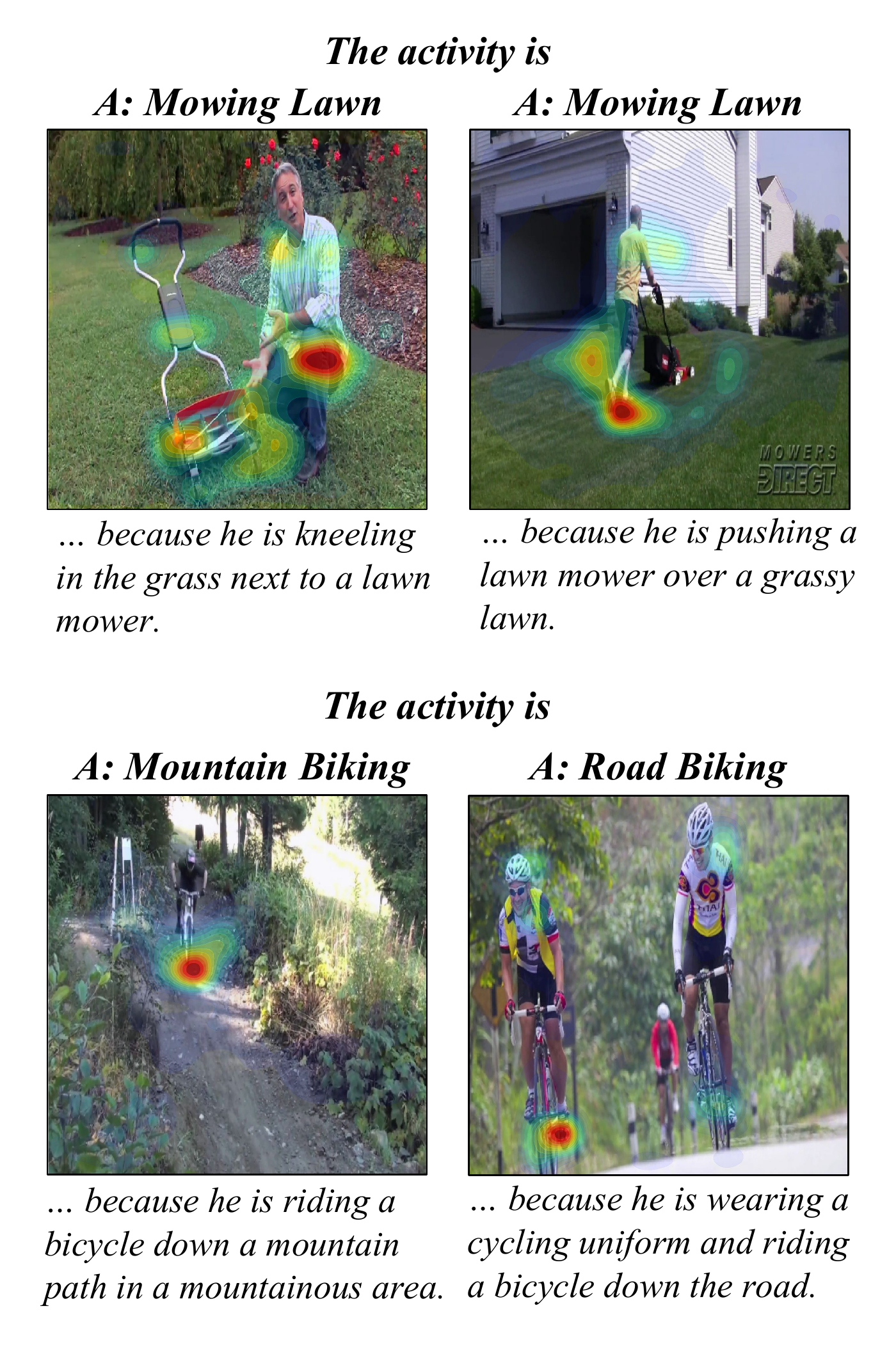}
\caption{ACT-X qualitative results: For each image the PJ-X model provides an answer and a justification, and points to the evidence for that justification.}
\label{fig:ACTqualitative}
\end{figure}

\subsection{Qualitative Results}
\label{sec:res:qual}
In this section we present our qualitative results on VQA-X and ACT-X datasets demonstrating that our model generates high quality sentences and the attention maps point to relevant locations in the image.

\myparagraph{VQA-X.}  \autoref{fig:VQAqualitative} shows qualitative results on our VQA-X dataset. We show pairs of images that form complementary pairs in VQA v2.
Our textual justifications are able to both capture common sense and discuss specific image parts important for answering a question.
For example, when asked ``Is this a zoo?'', the explanation model is able to discuss what the concept of ``zoo'' represents, \ie ``animals in an enclosure''.
When determining whether the water is calm, which requires discussing specific image regions, the textual justification discusses foam on the waves.

Visually, we notice that our attention model is able to point to important visual evidence.  
For example in the top row of~\autoref{fig:VQAqualitative}, for the question ``Is this a zoo?'' the visual explanation focuses on the field in one case, and the fence in another. 

\myparagraph{ACT-X.} \autoref{fig:ACTqualitative} shows results on our ACT-X dataset.
Textual explanations discuss a variety of visual cues important for correctly classifying activities such as 
global context, \eg ``a grassy lawn / a mountainous area'', and person-object interaction, \eg``pushing a lawn mower / riding a bicycle'' for mowing lawn and mountain biking, respectively.
These explanations require determining which of many multiple cues are appropriate to justify a particular action.

Our model points to visual evidence important for understanding each human activity.
For example to classify ``mowing lawn'' in the top row of ~\autoref{fig:ACTqualitative} the model focuses both on the person, who is on the grass, as well as the lawn mower. 
Our model can also differentiate between similar activities based on the context, \eg''mountain biking'' or ''road biking''.

\begin{figure}[t]
\center
\vspace{-0.2cm}
  \includegraphics[width=.8\linewidth]{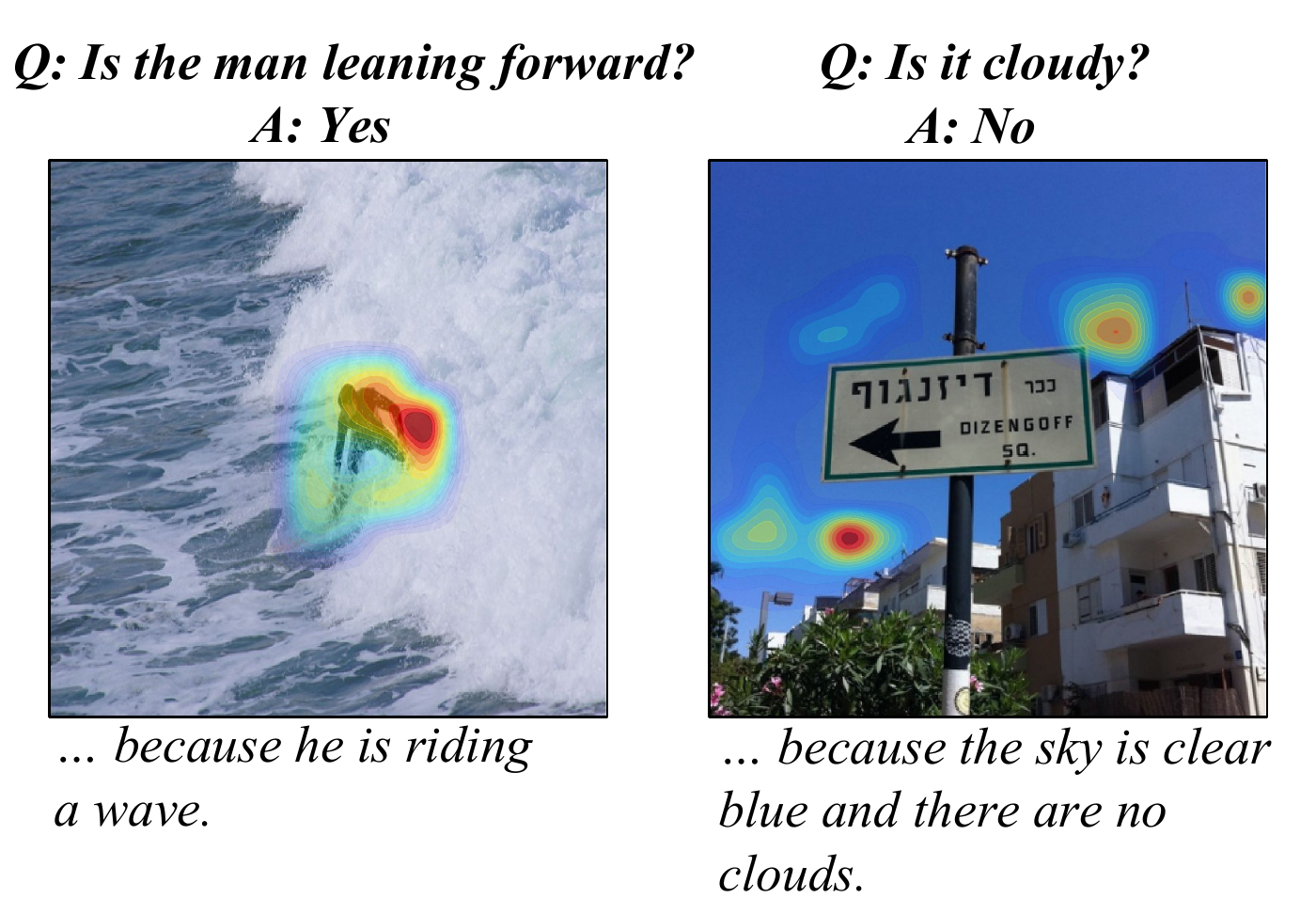}
  \vspace{-0.2cm}
\caption{Qualitative results comparing the insightfulness of visual pointing and textual justification. The left example demonstrates how visual pointing is more informative than textual justification whereas the right example shows the opposite.}
\label{fig:insightfulness}
\end{figure}

\myparagraph{Explanation Consistent with Incorrect Prediction.} Generating reasonable explnations for correct answers is important, but it is also crucial to see how a system behaves in the face of incorrect predictions. Such analysis would provide insights into whether the explanation generation component of the model is consistent with the answer prediction component or not. In~\autoref{fig:consistency}, we can see that the explanations are consistent with the incorrectly predicted answer for both VQA-X and ACT-X. For instance in the bottom-right example, we see that the model attends to a vacuum-like object and textually justifies the prediction ''vacuuming''. Such consistency between the answering model and the explanation model is also shown in \autoref{tbl:generation_automatic} where we see a drop in performance when explanations are conditioned on predictions (bottom rows) instead of the ground-truth answers (top rows).

\begin{figure}[t]
\center
  \includegraphics[width=.8\linewidth]{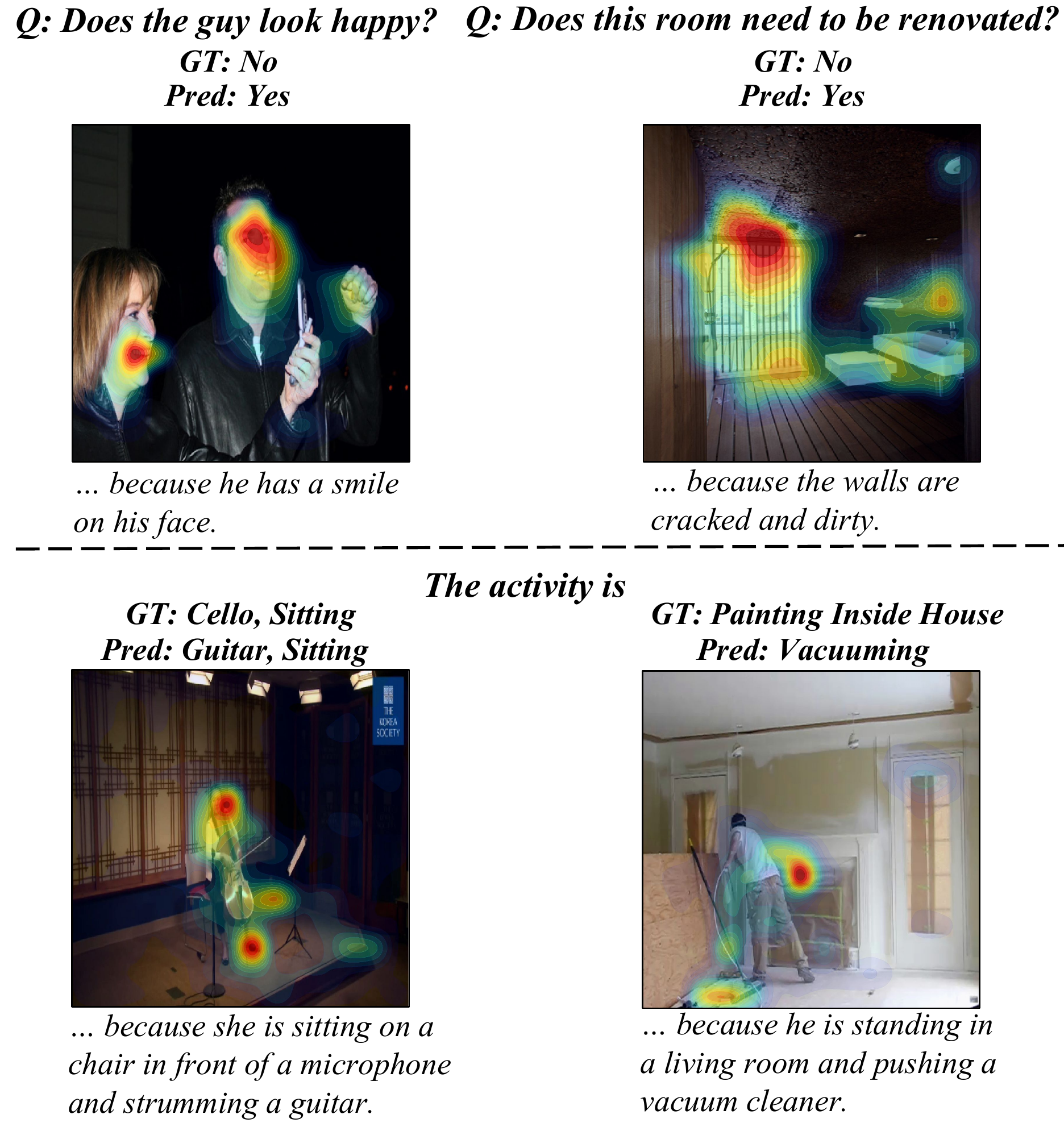}
\caption{Visual and textual explanations generated by our model conditioned on incorrect predictions.}
\label{fig:consistency}
\end{figure}

\subsection{Usefullness of Multimodal Explanations}
In this section, we address some of the advantages of generating multimodal explanations. In particular, we look at cases where visual explanations are more informative than the textual explanations, and vice versa. We also investigate how multimodal explanations can help humans diagnose the performance of an AI system.

\myparagraph{Complementary Explanations.} Multimodal explanations can support different tasks or support each other. %
Interestingly, in \autoref{fig:insightfulness}, we present some examples where visual pointing is more insightful than textual justification, and vice versa. Looking at the left example in \autoref{fig:insightfulness}, it is rather difficult to explain ``leaning'' with language and the model resorts to generating a correct, yet uninsightful sentence. However, the concept is easily conveyed when looking at the visual pointing result. In contrast, the right example shows the opposite. Looking at only some patches of the sky presented by the visual pointing result does not necessarily confirm if the scene is cloudy or not, while it is also unclear if attending to the entire region of the sky is a desired behavior. Yet, the textual justification succinctly captures the rationale. These examples clearly demonstrate the value of generating multimodal explanations.

\myparagraph{Diagnostic Explanations.} We evaluate an auxiliary task where humans have to guess whether the system correctly or incorrectly answered the question. The predicted answer is not shown; only image, question, correct answer, and textual/visual explanations. The set contains 50\% correctly answered questions. We compare our model against the models used for ablations in~\autoref{tbl:generation_automatic}. \autoref{tbl:human_pred_correctness} indicates that explanations are better than no explanations and our model is more helpful than models trained on descriptions and also models trained to generate textual explanations only.

\begin{table}[t]
\begin{center}
\begin{tabular}{l l l}
\toprule
& \textsc{vqa-x} & \textsc{act-x} \\
\midrule
Without explanation & 57.5\% & 51.5\% \\
Ours on Descriptions & 66.5\% &72.5\% \\
Ours w/o Attention & 61.5\% & 76.5\% \\
Ours & \textbf{70.0\%} &\textbf{80.5\%}\\
\toprule
\end{tabular}
\vspace{-2mm}
\vspace{-1mm}
\caption{Accuracy of humans judging if the model predicted correctly.}
\label{tbl:human_pred_correctness}
\end{center}
\end{table}
\section{Conclusion}
\label{sec:conc}
As a step towards explainable AI models, we proposed multimodal explanations for real-world tasks. Our model is the first to be capable of providing natural language justifications of decisions as well as pointing to the evidence in an image. We have collected two novel explanation datasets through crowd sourcing for visual question answering and activity recognition, \ie VQA-X and ACT-X. We quantitatively demonstrated that learning to point helps achieve high quality textual explanations. We also quantitatively show that using reference textual explanations to train our model helps achieve better visual pointing. 
Furthermore, we qualitatively demonstrated that our model is able to point to the evidence as well as to give natural sentence justifications, similar to ones humans give. %

\newpage

\clearpage
\bibliographystyle{ieee}
\bibliography{biblioLong,biblio,egbib}

\end{document}